%% file: sample-base.tex
\documentclass[sigconf,table]{acmart}
\usepackage{colortbl}
\usepackage{placeins}
\usepackage{float}

\usepackage{amssymb}
\usepackage{multicol}
\usepackage{balance}
\usepackage{lipsum}
\usepackage{hyperref}
\usepackage{multirow}
\usepackage{pifont}
\usepackage{footnote}
\usepackage{enumitem}
\usepackage[ruled,vlined,linesnumbered]{algorithm2e}
\usepackage{listings}

\AtBeginDocument{%
  }
\setcopyright{acmlicensed}
\copyrightyear{2025}
\acmYear{2025}
\acmDOI{XXXXXXX.XXXXXXX}
\acmConference[ACMMM25]{ACM International Conference on Multimedia}{October 27--31, 2025}{Dublin, Ireland}
\acmISBN{978-1-4503-XXXX-X/2018/06}

\begin{document}

\title{CheXPO: Preference Optimization for Chest X-ray VLMs \\ with Counterfactual Rationale}

\author{
    Xiao Liang\textsuperscript{1},
    Jiawei Hu\textsuperscript{1},
    Di Wang\textsuperscript{1,*},
    Zhi Ma\textsuperscript{1,*} ,
    Lin Zhao\textsuperscript{2},
    Ronghan Li\textsuperscript{1},
    Bo Wan\textsuperscript{1},
    Quan Wang\textsuperscript{1}
}

\affiliation{
    \textsuperscript{1} The Key Laboratory of Smart Human-Computer Interaction \\ and Wearable Technology of Shaanxi Province, Xidian University, Xi'an, China \\
    \textsuperscript{2} Nanjing University of Science and Technology, Nanjing, China
    \country{}
}


%

\renewcommand{\shortauthors}{Xiao Liang et al.}

\input{sec/0_abstract}


\begin{teaserfigure}
  \vspace{-0.2cm}
  \includegraphics[width=1\textwidth]{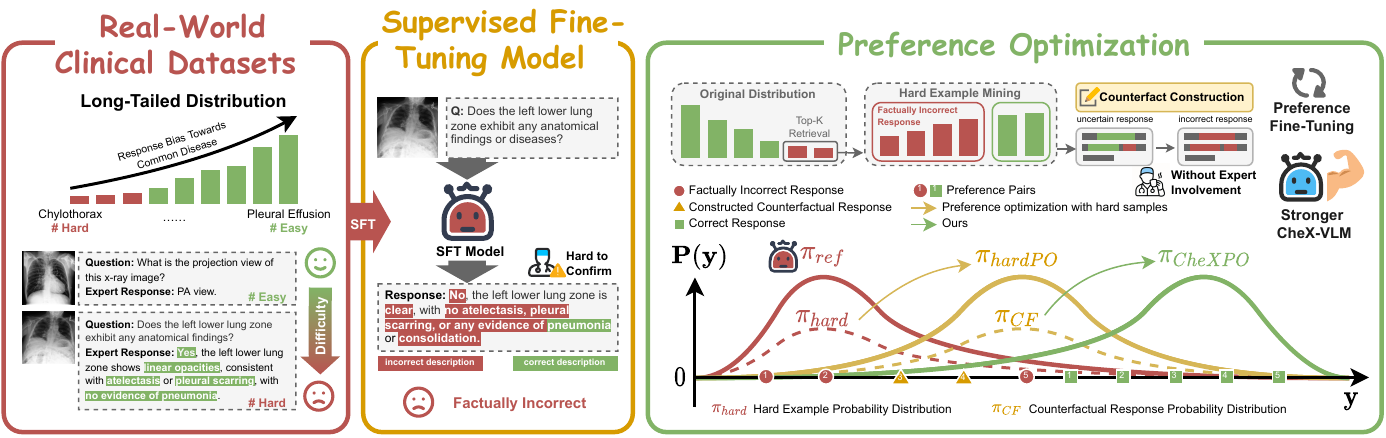}
  \vspace{-0.5cm}
  \caption{Naively adopting preference alignment for real-world clinical data makes it difficult to handle imbalanced distributions and challenging cases. Our \textit{CheXPO} employs preference optimization through hard example mining to balance the data distribution and automatically constructs counterfactuals as rejection responses to deal with challenging cases, resulting in more reliable outputs.}
  \label{fig:teaser}
\end{teaserfigure}


\maketitle
\begingroup\renewcommand\thefootnote{*}
\footnotetext{Corresponding authors.}
\endgroup

\input{sec/1_intro2}
\input{sec/2_relatedwork}

\input{sec/3.0_dataset}

\input{sec/3.1_methodology}

\input{sec/4_experimental}

\input{sec/5_result}
\input{sec/6_conclu}

\clearpage
\bibliographystyle{ACM-Reference-Format}
\bibliography{sample-base}

\clearpage
\appendix
\setcounter{section}{0}
\input{sec/7_appendix}

\end{document}

%% file: sec/0_abstract.tex
\begin{abstract}

Vision-language models (VLMs) are prone to hallucinations that critically compromise reliability in medical applications. While preference optimization can mitigate these hallucinations through clinical feedback, its implementation faces challenges such as clinically irrelevant training samples, imbalanced data distributions, and prohibitive expert annotation costs. To address these challenges, we introduce \textbf{CheXPO}, a \textbf{Che}st \textbf{X}-ray \textbf{P}reference \textbf{O}ptimization strategy that combines confidence-similarity joint mining with counterfactual rationale. Our approach begins by synthesizing a unified, fine-grained multi-task chest X-ray visual instruction dataset across different question types for supervised fine-tuning (SFT). We then identify hard examples through token-level confidence analysis of SFT failures and use similarity-based retrieval to expand hard examples for balancing preference sample distributions, while synthetic counterfactual rationales provide fine-grained clinical preferences, eliminating the need for additional expert input. Experiments show that CheXPO achieves 8.93\% relative performance gain using only 5\% of SFT samples, reaching state-of-the-art performance across diverse clinical tasks and providing a scalable, interpretable solution for real-world radiology applications. The code and dataset are available here \footnote{\url{https://github.com/ResearchGroup-MedVLLM/CheX-Phi35V}}.

\end{abstract}

%% file: sec/1_intro2.tex
\section{Introduction}
\label{sec:intro}

\balance
Vision-language models (VLMs) \cite{openai2024gpt4o, llava, qwen} have driven advances in open-domain dialogue and understanding across various domains. In the medical field, numerous approaches \cite{llava-med, pmlr-v225-moor23aMed-Flamingo, DBLP:journals/corr/abs-2308-02463_RadFM, Omkar2023XrayGPT, wang2023XrayGLM, chen2024huatuogptvisioninjectingmedicalvisual, Cui2024BiomedicalVI} have emerged, aimed at tasks such as image interpretation, pathology analysis, and medical question answering. These methods build large-scale medical visual instruction datasets and employ supervised fine-tuning (SFT) to adapt general-purpose VLMs to medical tasks. However, these medical vision-language models still face reliability issues, particularly factual errors, also known as ``\textit{hallucinations},” where models generate responses that appear coherent but contradict the image information.

Existing research \cite{sun2024stllava, hein2024preference} attributes the primary cause of hallucinations in MedVLMs to modality misalignment \cite{zhou2024aligningmodalitiesvisionlarge} and attempts to improve alignment through constructing preference data to optimize models. However, the preference data generation processes for these methods largely follow approaches from the general text-based \cite{Rafailov2023DirectPO, Gao2025PrincipledDS, Morimura2024FilteredDP} or vision-language domains \cite{Liu2025VisualRFTVR, OPA-DPOYang2025MitigatingHI, Liu2024MIADPOMA}, overlooking the unique characteristics and requirements of real clinical scenarios. Specifically, the following challenges exist: \textbf{\ding{172} Training samples lack clinical significance.} Questions such as ``\textit{Is this image an AP view or a PA view?}'' can be easily addressed by an SFT model and offer limited value for clinical diagnostics, diminishing their usefulness for preference alignment. \textbf{\ding{173} Long-tailed distribution in real clinical data}. Medical datasets often exhibit imbalances between common diseases (e.g., ``\textit{Pleural Effusion}'') and rare diseases (e.g., ``\textit{Chylothorax}''), which leads to biased model outputs that favor more common diseases \cite{li2023dynamicCL, Li2024ContrastiveLW}, undermining the model's ability to accurately diagnose rare diseases. \textbf{\ding{174} High cost of preference data creation}. Generating preference data for medical image responses requires expert evaluation, making the process costly and limiting scalability, which hinders further improvements in model performance.

In this paper, we introduce \textbf{CheXPO}, a \textbf{P}reference \textbf{O}ptimization strategy for \textbf{Che}st \textbf{X}-ray vision-language models, designed to mitigate the aforementioned issues through confidence-similarity joint hard example mining and counterfactual rationale, as shown in Figure \ref{fig:teaser}. Specifically, we unify multiple benchmarks \cite{Johnson2019_MIMICCXRAD, Bae2023EHRXQAAM, Hu2023ExpertKI_MedDiffVQA, Wu2021ChestID_ChestImaGenome} into a large-scale visual instruction dataset containing 3 specialized question-answering tasks (Basic-QA, Region-QA, and Comparison-QA), 10 question types, and 640k expert-level rationales, enabling general VLMs to interpret chest X-rays through SFT. The proposed CheXPO enhances the reliability of SFT models through three steps: \textbf{\ding{172}} Token-level confidence analysis on SFT models identifies clinically significant responses, including errors and low-confidence responses as hard examples. \textbf{\ding{173}} Similarity-based hard example expansion with BioMedCLIP \cite{Zhang2023BiomedCLIPAM}, retrieving semantically similar samples to balance the long-tailed distributions of the preference dataset. \textbf{\ding{174}} Constructing counterfactual rationale as rejection responses, leveraging preference optimization to reinforce fine-grained supervision, particularly for misjudgment errors related to abnormality, anatomy, and severity. Our proposed CheXPO demonstrates strong scalability, eliminating the need for expert annotations or GPT API dependencies, while maintaining cost-effectiveness and implementation efficiency for large-scale clinical applications. Our contributions are summarized as follows:
\begin{itemize}
    \item A large-scale synthetic chest X-ray vision-language instruction dataset covering basic, region, and comparison tasks, with 10 clinically fine-grained question types and 640k expert-level rationales for supervised fine-tuning.
    \item A scalable preference optimization strategy, CheXPO, utilizes data from only 5\% of the SFT samples through confidence-similarity-joint hard example mining and counterfactual rationale, achieving an 8.93\% relative improvement in performance without additional expert annotations.
    \item Our model, \textbf{CheX-Phi3.5V}, outperforms existing methods on the MIMIC-CXR-VQA and Medical-Diff-VQA benchmarks, demonstrating improved interpretability and reliability, thereby better meeting practical clinical needs.
\end{itemize}

%% file: sec/2_relatedwork.tex
\section{Related Work}
\label{sec:related}

\subsection{Medical Vision-Language Models}
The development of vision-language models (VLMs) has significantly advanced task-specific applications in the medical field. Using GPT-based instruction generation, supervised fine-tuning (SFT) has become the primary method for adapting general-purpose VLMs to meet medical needs. LLaVA-Med \cite{llava-med} employs alignment and instruction tuning to enable open-ended question answering for biomedical images. Med-Flamingo \cite{pmlr-v225-moor23aMed-Flamingo} enhances few-shot reasoning through a multimodal context learning framework. RadFM \cite{DBLP:journals/corr/abs-2308-02463_RadFM} integrates large-scale radiology datasets for more robust multi-task image analysis. XrayGPT \cite{Omkar2023XrayGPT} and XrayGLM \cite{wang2023XrayGLM} focus on multi-turn QA and automated chest X-ray interpretation. HuatuoVision \cite{chen2024huatuogptvisioninjectingmedicalvisual} generates 1.3 million VQA instances to train high-quality question-answering models. BioMed-VITAL \cite{Cui2024BiomedicalVI} integrates expert-selected data, which notably enhances its performance in complex clinical scenarios. As VLMs continue to advance, scaling synthetic SFT data broadens their scenario coverage and enhances MedVLM robustness. However, in real-world clinical settings, the quality of medical datasets—such as long-tail distributions, unverifiable missing data, or erroneous entries—poses significant challenges to the reliability of MedVLMs. Moreover, SFT's focus on next-token prediction—a task inherently biased toward memorizing training patterns—limits its ability to generalize, thereby constraining adaptability to the complex and evolving demands of clinical practice.

\subsection{Human Feedback in VLM Alignment}
To overcome SFT limitations and improve model reliability, reinforcement learning from human feedback (RLHF) has gained prominence. Direct Preference Optimization (DPO) \cite{Rafailov2023DirectPO} offers a simpler, cost-effective alternative, widely used to enhance model alignment with preferences. DPO relies heavily on well-chosen preference pairs, which, if noisy, risk suboptimal policies or overfitting \cite{DBLP:journals/corr/abs-2502-14560}. Recent approaches address these challenges through improved data selection and filtering. Selective DPO \cite{Gao2025PrincipledDS} employs margin-based filtering for higher-quality pairs, while Filtered DPO \cite{Morimura2024FilteredDP} excludes unreliable samples. MIA-DPO \cite{Liu2024MIADPOMA} uses attention mechanisms to select rejected samples and reduce noise. CLIP-DPO \cite{Ouali2024CLIPDPOVM} filters mismatched pairs via CLIP, enhancing preference data reliability. In medical applications, STLLaVA-Med \cite{sun2024stllava} combines self-supervised generation and DPO fine-tuning to align with radiologist preferences. CheXalign \cite{hein2024preference} uses reference-based alignment, while RULE \cite{DBLP:conf/emnlp/XiaZLZLLZY24} and MMed-RAG employ retrieval-based strategies to improve preference selection. MMedPO \cite{Zhu2024MMedPOAM} applies multimodal optimization, reducing hallucinations and improving diagnostic accuracy. While existing methods enhance VLM factual consistency and hallucination mitigation through modality/data-source adaptations, they inadequately address practical chest X-ray diagnostic challenges (e.g., fine-grained pathology analysis) and suffer from low data efficiency.


%% file: sec/3.0_dataset.tex
\section{Preliminary: Dataset Synthesis}
\label{sec:dataset}

To build a comprehensive chest X-ray visual instruction dataset, we combine four sources: MIMIC-CXR \cite{Johnson2019_MIMICCXRAD} for images and reports; Chest ImaGenome \cite{Wu2021ChestID_ChestImaGenome} for expert-labeled organs and diseases; and MIMIC-CXR-VQA \cite{Bae2023EHRXQAAM} and Medical-Diff-VQA \cite{Hu2023ExpertKI_MedDiffVQA} for all image QA pairs. Despite the extensive clinical questions provided by \cite{Bae2023EHRXQAAM} and \cite{Hu2023ExpertKI_MedDiffVQA}, their short answers are insufficient to equip vision-language models with dialogue and rationale generation capabilities. Given that these datasets share chest X-rays, study IDs, and subject IDs, MIMIC-CXR provides radiologist reports, supporting fine-grained medical reasoning and structured diagnostic interpretations. Therefore, we leverage the powerful GPT-4o \cite{openai2024gpt4o}, synthesizing existing questions/answers with novel GPT-4o-generated rationales and clinical attributes to simulate expert diagnostic processes, significantly reducing manual annotation costs. Below is a detailed description of each task:

\textbf{Basic-QA}. We utilize a structured prompting template that integrates existing question-answer pairs from MIMIC-CXR-VQA \cite{Bae2023EHRXQAAM} and corresponding radiology reports from MIMIC-CXR \cite{Johnson2019_MIMICCXRAD}. The model is instructed to reference these reports and infer supporting evidence from the image to justify the given answer. It should be emphasized that the rationale \(\mathcal{T}\) synthesized by GPT-4o must begin with a short answer \(\mathcal{A}\). The short answer can either be a single word, such as "\textit{Yes}," or a phrase composed of multiple words, such as "\textit{Left lung and right lung}." The explanation \(\mathcal{E}\) must be provided from the perspective of an expert reviewing the image, answering the question \(\mathcal{Q}\), and supporting the gold standard answers provided in MIMIC-CXR-VQA.

\textbf{Region-QA}. To enhance interpretability, we build on existing question-answer pairs by adding explicit visual prompts via alpha-channel markers, and we replace specific anatomical terms in the question with placeholders such as \textit{"marked region"} or \textit{"highlighted area."} The visual prompts refer to the bounding box coordinates provided by the Chest ImaGenome \cite{Wu2021ChestID_ChestImaGenome} silver dataset. To further enhance spatial flexibility, we additionally incorporate prompt types such as arrows, ellipses, or asterisks (*), following the approach of \cite{Cai2023ViPLLaVAML}. The arrowheads are positioned within a constrained region $[(-w/2, -h/2),$ $ (w/2, h/2)]$, where \(w\) and \(h\) represent the bounding box width and height, respectively. For ellipses, the axes inherit dimensions from the bounding box and are scaled within a range of \([1, 1.5]\). These augmentations are inspired by visual prompts commonly seen in PubMed figures, where experts directly overlay markers on images for clearer illustration, providing a more intuitive and flexible approach compared to \cite{Mller2024ChEXIL}.

\textbf{Comparison-QA}. Medical-Diff-VQA\cite{Hu2023ExpertKI_MedDiffVQA} provides study IDs for both the main and reference chest X-rays being compared. We leverage these IDs to retrieve their respective radiology reports, using them as contrastive evidence. GPT-4o is then prompted to generate detailed comparative explanations based on the main report and the prior report. Similar to Basic and Region-QA, we instruct GPT-4o to emulate expert image analysis, requiring synthesized rationales to derive exclusively from visual evidence while strictly avoiding references to textual reports or prior diagnostic annotations. We unify the question types from MIMIC-Ext-VQA and Med-Diff-VQA into 10 clinically relevant categories, including Abnormality, Presence, Location, View, Level, Type, Difference, Size, Gender, and Attribute.

\begin{figure}[t]
    \centering
    \includegraphics[width=\columnwidth]{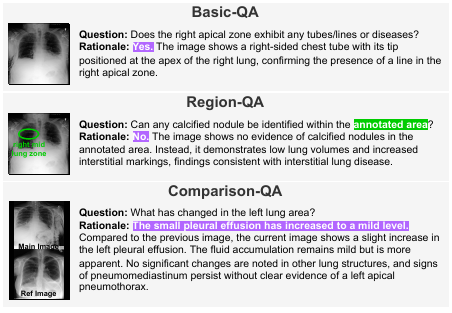}
    \setlength{\fboxsep}{1pt} 
    \vspace{-0.5cm}
    \caption{
    \textbf{Example of integrated multi-task QA dataset.} Each rationale $\mathcal{T}$ includes a  {\colorbox[HTML]{B266FF}{\textcolor{white}{short answer $\mathcal{A}$}}} and a GPT-4o-generated {\colorbox[HTML]{DBDBDB}{explanation $\mathcal{E}$}}. The  {\colorbox[HTML]{00CC00}{\textcolor{white}{"annotation area"}}} indicates the visual prompt on the image.
    }
    
    \label{fig:multi_task_qa}
\end{figure}

Figure \ref{fig:multi_task_qa} illustrates examples of the three tasks. To ensure fair benchmarking, we follow the original train/validation/test splits, with additional cleaning to address missing MIMIC-CXR reports, absent Chest ImaGenome bounding boxes, and overlapping images across benchmarks. Detailed descriptions of each data source and final dataset statistics, along with detailed counts and rationale generation prompts, can be found in the Appendix. All the above datasets are publicly available on the PhysioNet platform \cite{PhysioNet} and require credentialed access approval.

%% file: sec/3.1_methodology.tex
\begin{figure*}[t]
    \centering
    \includegraphics[width=0.9\textwidth]{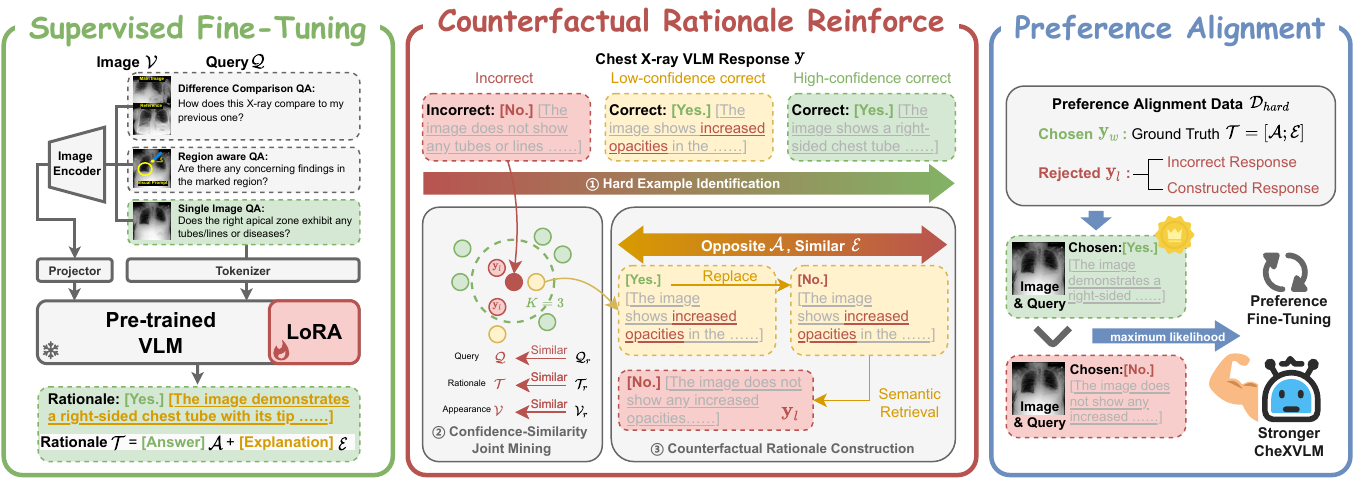}
    \caption{\textbf{Overview of our chest X-ray VLM training pipeline.}
    Our framework leverages visual instruction tuning and preference alignment, fine-tuning Phi-3.5V with LoRA to obtain a base SFT model.
    A confidence-similarity joint mining strategy selects high-value samples, followed by counterfactual rationale construction for preference alignment to reinforce clinical reliability.}

    \label{fig:method_overview}
\end{figure*}

\section{Methodology}
\label{sec:method}

In this section, we introduce our training pipeline, as shown in Figure~\ref{fig:method_overview}. We first fine-tune Phi-3.5V \cite{Abdin2024Phi3TR} with the synthetic instruction dataset to obtain a supervised base model, CheX-Phi3.5V. The \textbf{CheXPO} strategy is then employed, utilizing confidence-similarity joint mining to select hard examples, followed by counterfactual rationale construction to obtain preference pairs for preference alignment. The complete method workflow is detailed in Algorithm~\ref{alg:stratified_sampling}. The following sections describe each component in detail. To ensure reproducibility, we provide implementation and hyperparameter details in the Appendix.

\subsection{Hard Example Identification}

In the training pipeline of our CheX-Phi3.5V, SFT primarily teaches the model to understand medical images and follow instructions to generate answers in the desired format, while direct preference optimization (DPO) \cite{Rafailov2023DirectPO} further refines the model's clinical reliability and response preference. However, imbalanced data is common in chest X-ray datasets, which tend to overrepresent healthy cases or frequent findings \cite{Mller2024ChEXIL, li2023dynamicCL, Li2024ContrastiveLW, wang2023metr}. This imbalance can lead the SFT model to overfit to common diagnostic patterns, while DPO risks exacerbating distributional shifts by disproportionately reinforcing majority-class patterns through biased preference pairs. To address this issue, we introduce a hard example mining strategy that identifies high-value preference data for DPO training, specifically targeting cases where the base SFT model fails. Given a chest X-ray vision-language model \(\mathcal{M}(.)\) trained via supervised fine-tuning, and an SFT training dataset \(\mathcal{D}\) consisting of images \(\mathcal{V}\), questions \(\mathcal{Q}\), and rationales \(\mathcal{T}\), where \(\mathcal{T}\) is a concise answer followed by a detailed explanation, formally represented as \(\mathcal{T} = [\mathcal{A}; \mathcal{E}]\). To efficiently evaluate model performance and identify hard examples, we perform stratified random sampling from the training dataset based on question and answer types and obtain a representative subset \(\mathcal{D}_{sample}\) for initial assessment. The stratified sampling ratio, denoted as \(\gamma\), represents the proportion of the dataset selected, with \(\gamma < 5\%\) of the total training data, ensuring comprehensive coverage of all question types. Within this subset, for each image $\mathcal{V}$ and corresponding question $\mathcal{Q}$, the model $\mathcal{M}$ generates a token sequence $\mathbf{y} = [\mathbf{y}_\mathcal{A}; \mathbf{y}_\mathcal{E}]$ through auto-regressive factorization. As only the short answer segments $\mathbf{y}_\mathcal{A}$ have well-defined ground truth labels, we focus on them and directly compute their log-probabilities. To prevent longer answers from being unfairly penalized due to their length, we apply a length-normalization technique inspired by \cite{Malinin2021UncertaintyEI}:

\begin{equation}
\label{eq:length_norm_logprob}
\mathbf{p} = \log \tilde{\mathbf{P}}( \mathbf{y}_\mathcal{A} | \mathcal{V}, \mathcal{Q}; \theta) = \frac{1}{T_\mathcal{A}} \sum_{t=1}^{T_\mathcal{A}} \log \mathbf{P}(y_t | y_{<t}, \mathcal{V}, \mathcal{Q}; \theta),
\end{equation} where $y_{<t} \triangleq [y_1, ..., y_{t-1}]$ represents the prefix sequence, and $T_\mathcal{A}$ denotes the length of the short answer sequence. When $\mathbf{y}_\mathcal{A} \neq \mathcal{A}$, the predicted answer $\mathbf{y}_\mathcal{A}$ is directly labeled as a rejected response $\mathbf{y}_l$ and included in a new subset $\mathcal{D}_{hard}$ for DPO training, reinforcing factual grounding and reducing confident errors. Conversely, when $\mathbf{p}$ is low (i.e., $\mathbf{p} < \sigma$, a predefined threshold) yet $\mathbf{y}_\mathcal{A} = \mathcal{A}$, the model is likely uncertain about its prediction. In this case, we construct contrastive responses as the rejected $\mathbf{y}_l$ based on expert knowledge, enhancing model differentiation through DPO training.

\subsection{Confidence-Similarity Joint Mining}
After initializing the hard example set \(\mathcal{D}_{hard}\), where \(\mathcal{D}\) consists of images \(\mathcal{V}\), questions \(\mathcal{Q}\), and rationales \(\mathcal{T}\), formally represented as \(\mathcal{D} = \{(\mathcal{V}^{(i)}, \mathcal{Q}^{(i)}, \mathcal{T}^{(i)})\}_{i=1}^N\), we retrieve semantically similar samples from the remaining data \(\mathcal{D}_{rest} = \mathcal{D} - \mathcal{D}_{hard}\) to further enrich the dataset for DPO training, thereby mitigating SFT failures caused by the long-tailed distribution. We first leverage BioMedCLIP ~\cite{Zhang2023BiomedCLIPAM} to extract features $f_\mathcal{V}$, $f_\mathcal{Q}$, and $f_\mathcal{T}$ for the images, questions, and rationale in $\mathcal{D}$. Then, we compute cosine similarity matrices $\mathcal{S}$ between $\mathcal{D}_{hard}$ and $\mathcal{D}_{rest}$ for $f_\mathcal{Q}$, $f_\mathcal{T}$, and $f_\mathcal{V}$, respectively, to identify samples with similar semantics and visual context:

\begin{equation}
    \mathcal{S}_\mathcal{X} = \Big[ \langle f_\mathcal{X}^{(i)}, f_\mathcal{X}^{(j)} \rangle \Big]_{\substack{i \in \mathcal{D}_{hard} \\ j \in \mathcal{D}_{rest}}}, \quad \mathcal{X} \in \{\mathcal{Q}, \mathcal{T}, \mathcal{V}\},
\end{equation} where $\langle \cdot, \cdot \rangle$ denotes cosine similarity. Final similarity combines all modalities: 
\begin{equation}
\label{eq:combined_simi}
\mathcal{S} = \mathcal{S}_\mathcal{Q} + \mathcal{S}_\mathcal{T} + \mathcal{S}_\mathcal{V},
\end{equation} where $\mathcal{S} = \{\mathbf{s}^{(i,j)}\}_{i \in \mathcal{D}_{hard}, j \in \mathcal{D}_{rest}}$. We then select the Top-$K$ most similar samples $\mathcal{N}^{(i)}$ for $i$-th hard example based on $\mathcal{S}$:
\begin{equation}
\label{eq:topk}
\mathcal{N}^{(i)} = \text{TopK}_{j \in \mathcal{D}_{rest}} \, \mathcal{S}^{(i,j)}, \quad \forall i \in \mathcal{D}_{hard},
\end{equation} and this combined similarity ensures a balance between semantic alignment and visual consistency, as further validated in the ablation study (Section~\ref{sec:abl}).



\subsection{Counterfactual Rationale Construction}

While incorrect samples are directly used as rejected responses, correct predictions (i.e., $\mathbf{y}_\mathcal{A} = \mathcal{A}$) are underutilized due to unverifiable explanatory rationales $\mathcal{E}$, limiting data efficiency. Given that low-confidence predictions are prone to errors \cite{Factual_Confidence, DBLP:conf/emnlp/XiaZLZLLZY24}, we analyze samples with log-probability scores $\mathbf{p}$ below the threshold $\sigma$ and find that \texttt{Abnormality}, \texttt{Anatomy}, and \texttt{Severity} misjudgments collectively account for over 70\% of model errors (see Section~\ref{sec:abl}). In response, we propose a chest X-ray-specific counterfactual rationale construction strategy by designing domain-specific rejection pools:
\begin{itemize}
    \item \textbf{Anatomy Rejection Pool:} Contains semantically similar but clinically opposite anatomical terms (e.g., ``\textit{left lung opacity}'' vs. ``\textit{right lung opacity}'') to generate counterfactual rationale.
    \item \textbf{Abnormality Rejection Pool:} Includes confusable disease conditions (e.g., ``\textit{pneumonia}'' vs. ``\textit{pulmonary edema, atelectasis}'' vs. ``\textit{pneumothorax}'') to contrast common misdiagnoses.
    \item \textbf{Severity Rejection Pool:} Modifies severity indicators (e.g., ``\textit{mild interstitial infiltrates}'' vs. ``\textit{severe consolidation}'', ``\textit{acute phase}'' vs. ``\textit{chronic stage}'') to introduce graded diagnostic contradictions.
\end{itemize}

Specifically, for MedVLM-generated $\mathbf{y} = [\mathbf{y}_\mathcal{A}; \mathbf{y}_\mathcal{E}]$, we first classify $\mathbf{y}_\mathcal{A}$'s type and randomly replace it using the corresponding rejection pool, obtaining $[\overline{\mathbf{y}_\mathcal{A}}; \mathbf{y}_\mathcal{E}]$. For question types such as \texttt{Gender} and \texttt{View}, where answers have clear opposites (e.g., ``\textit{female}'' vs. ``\textit{male}'', ``\textit{AP view}'' vs. ``\textit{PA view}''), we directly use the opposite answer as $\overline{\mathbf{y}_\mathcal{A}}$. For types like \texttt{Size} and \texttt{Type}, we randomly sample a different answer of the same type as a contrastive replacement. We then retrieve the Top-1 contrastive candidate $\overline{\mathbf{y}}$ based on BioMedCLIP:

\begin{equation}
\label{eq:replace}
\overline{\mathbf{y}} = \underset{\mathcal{T}_k \in \mathcal{D}_{rest}}{\arg\max} \ \langle f_{\mathcal{T}_k}, f_{[\overline{\mathbf{y}_\mathcal{A}}; \mathbf{y}_\mathcal{E}]} \rangle. 
\end{equation}

Subsequently, we assign $\overline{\mathbf{y}}$ as $\mathbf{y}_l$, serving as the rejected response in the preference pair used for alignment. This ensures rejected responses maintain semantic coherence while presenting clinically counterfactual rationale.

\input{sec/3.2_algorithm}

\subsection{Preference Alignment with DPO}

Direct Preference Optimization (DPO) \cite{Rafailov2023DirectPO} aligns language models with human preferences by reformulating reward optimization as supervised learning. It treats the base model as a reference policy and optimizes a new policy to favor preferred responses over dispreferred ones using a pairwise logistic loss:
\begin{equation}
\label{eq1:dpo}
\begin{aligned}
\mathcal{L}_{\mathrm{DPO}}(\pi_{\boldsymbol{\theta}}, \mathcal{D}) =& -\mathbb{E}_{(x,\mathbf{y}_w,\mathbf{y}_l \sim \mathcal{D})} \Big[ \Big. \\
&\log \phi ( \beta \log \frac{\pi_{\boldsymbol{\theta}}(\mathbf{y}_w|x)}{\pi_{ref}(\mathbf{y}_w|x)} - \beta \log \frac{\pi_{\boldsymbol{\theta}}(\mathbf{y}_l|x)}{\pi_{ref}(\mathbf{y}_l|x)} ) \Big. \Big].
\end{aligned}
\end{equation} where $\mathbf{y}_w$ is a human-preferred response and $\mathbf{y}_l$ is a less preferred response, $\pi_\theta$ is the tuned model, and $\pi_{ref}$ is the reference model. $\phi$ is the sigmoid function, and $\beta$ is a temperature hyperparameter. Intuitively, DPO pushes $\pi_\theta$ to assign a higher score to the preferred output than to the rejected one while constraining changes relative to the reference model. It can be easily extended to optimizing our chest X-ray vision-language model, using preference-aligned pairs obtained via the proposed confidence-similarity joint mining and counterfactual rationale construction strategies. Through the proposed method, we efficiently leverage existing data to maximize the potential of priors in chest X-rays, utilizing domain-specific knowledge as implicit preference signals. The complete data construction and training pipeline is detailed in Algorithm~\ref{alg:stratified_sampling}.


%% file: sec/3.2_algorithm.tex
\begin{algorithm}[htbp]
\caption{CheXPO Strategy Pipeline}
\label{alg:stratified_sampling}

\KwIn{
  Dataset \(\mathcal{D}=\{(\mathcal{V},\mathcal{Q},\mathcal{T}=[\mathcal{A};\mathcal{E}])\}_{i=1}^{N}\); \\
  Sampled subset \(\mathcal{D}_{sampled}\); 
  Med-LVLM \(\mathcal{M}(\cdot,\cdot)\)
}
\KwOut{Preference alignment dataset \(\mathcal{D}_{hard}\).}

Initialize \(\mathcal{D}_{hard} \gets \emptyset\)\;

\ForEach{\((\mathcal{V},\mathcal{Q},\mathcal{T}) \in \mathcal{D}_{sampled}\)}{
    \(\mathbf{y} = [\mathbf{y}_\mathcal{A}; \mathbf{y}_\mathcal{E}] \gets \mathcal{M}(\mathcal{V},\mathcal{Q})\)\;
    \If{\(\mathbf{y}_\mathcal{A} \neq \mathcal{A}\)}{  

        \(\mathbf{y}_w \gets \mathcal{T}\), \(\mathbf{y}_l \gets \mathbf{y}\)\;
        Add \((\mathcal{V},\mathcal{Q},\mathbf{y}_w,\mathbf{y}_l)\) to \(\mathcal{D}_{hard}\)\;
    }
}

\ForEach{\((\mathcal{V},\mathcal{Q},\mathcal{T}) \in \mathcal{D}_{hard}\)}{
    Define \(\mathcal{D}_{rest} = \mathcal{D} \setminus \mathcal{D}_{hard}\)\;
    Retrieve TopK similar samples $\mathcal{N}$ from $\mathcal{D}_{rest}$ (Eq.~\ref{eq:topk})\;
    \ForEach{\((\mathcal{V},\mathcal{Q},\mathcal{T})\) $\in \mathcal{N}$}{
        \(\mathbf{y} = [\mathbf{y}_{\mathcal{A}};\mathbf{y}_{\mathcal{E}}] \gets \mathcal{M}(\mathcal{V},\mathcal{Q})\)\;
        Compute length-normalized log-prob \(\mathbf{p}\) (Eq.~\ref{eq:length_norm_logprob})\;
        
        \If{\(\mathbf{y}_{\mathcal{A}} \neq \mathcal{A}\)}{
            \(\mathbf{y}_w \gets \mathcal{T}\), \(\mathbf{y}_l \gets \mathbf{y}\)\;
        }
        \ElseIf{\(\mathbf{y}_{\mathcal{A}} = \mathcal{A} \land \mathbf{p} < \sigma\)}{
            \(\mathbf{y}_w \gets \mathcal{T}\), generate \(\mathbf{y}_l\) (Eq.~\ref{eq:replace})\;
        }
        Add \((\mathcal{V},\mathcal{Q},\mathbf{y}_w,\mathbf{y}_l)\) to \(\mathcal{D}_{hard}\)\;
    }
}

\end{algorithm}

%% file: sec/4_experimental.tex
\begin{table*}[t]
  \centering
  \resizebox{\linewidth}{!}{
    \begin{tabular}{lrrrrrrrrrrrrrr}
    \toprule
    \textbf{Model} & \multicolumn{10}{c}{\textbf{Question Type}}                                   &       & \multicolumn{2}{c}{\textbf{Answer Type}} & \multicolumn{1}{c}{\textbf{Overall}} \\
\cmidrule{2-11}\cmidrule{13-14}          & \multicolumn{1}{l}{Presence} & \multicolumn{1}{l}{Abnormality} & \multicolumn{1}{l}{Anatomy} & \multicolumn{1}{l}{Severity} & \multicolumn{1}{l}{Plane} & \multicolumn{1}{l}{Type} & \multicolumn{1}{l}{Difference} & \multicolumn{1}{l}{Attribute} & \multicolumn{1}{l}{Size} & \multicolumn{1}{l}{Gender} &       & \multicolumn{1}{l}{Open} & \multicolumn{1}{l}{Closed} &  \\
    \midrule
    \rowcolor[rgb]{ .851,  .851,  .851} \multicolumn{15}{l}{\textit{General Visual Language Models}} \\
    LLaVA-v1.6-7B \cite{llava} & 67.67 & 63.97 & 17.92 & 0.00  & 67.71 & 23.81 & 0.00  & 15.17 & 58.18 & 70.00 &       & 11.07 & 77.52 & 47.26 \\
    Qwen-VL-Chat \cite{qwen} & 66.95 & 63.02 & 21.67 & 18.81 & 88.95 & 14.29 & 81.25 & 21.38 & 45.45 & 36.67 &       & 25.91 & 76.31 & 53.36 \\
    Phi3.5V-4.2B \cite{Abdin2024Phi3TR} & 57.96 & 27.47 & 11.19 & 1.11  & 38.03 & 0.00  & 0.00  & 17.24 & 18.18 & 63.33 &       & 7.47  & 51.19 & 31.30 \\
    \midrule
    \rowcolor[rgb]{ .851,  .851,  .851} \multicolumn{15}{l}{\textit{Medical Visual Language Models}} \\
    Med-Flamingo \cite{pmlr-v225-moor23aMed-Flamingo} & 62.45 & 57.67 & 17.36 & 14.16 & 50.99 & 12.38 & 0.00  & 17.93 & 65.45 & 50.00 &       & 14.84 & 67.72 & 43.64 \\
    RadFM \cite{DBLP:journals/corr/abs-2308-02463_RadFM} & 59.51 & 48.53 & 12.36 & 7.08  & 40.79 & 8.57  & 0.00  & 13.79 & 63.64 & 63.33 &       & 9.75  & 62.06 & 38.24 \\
    LLaVA-Med-7B \cite{llava-med} & 68.21 & 64.22 & 17.92 & 0.00  & 68.56 & 27.14 & 0.00  & 16.55 & 45.45 & 70.00 &       & 11.51 & 77.78 & 47.60 \\
    HuatuoV-7B \cite{chen2024huatuogptvisioninjectingmedicalvisual} & 76.42 & 63.88 & 22.36 & 32.96 & 83.00 & 20.00 & 48.08 & 24.83 & 83.64 & 53.33 &       & 26.17 & 83.07 & 57.16 \\
    HuatuoV-34B \cite{chen2024huatuogptvisioninjectingmedicalvisual} & 90.82 & 61.55 & 22.92 & 54.87 & 89.24 & 14.76 & 0.00  & 27.59 & \textbf{92.73} & 66.67 &       & 27.27 & 90.97 & 61.96 \\
    \midrule
    \rowcolor[rgb]{ .851,  .851,  .851} \multicolumn{15}{l}{\textit{Baseline Implementation}} \\
    Phi3.5V+SA & 81.00 & 62.89 & 44.87 & 60.03 & 95.69 & 63.84 & 91.18 & 73.21 & 63.45 & 66.98 &       & 55.82 & 81.76 & 69.95 \\
    Phi3.5V+RA (Baseline) & 86.26 & 71.46 & 43.86 & 62.16 & 98.30 & 64.74 & 91.18 & 78.80 & 63.62 & 70.68 &       & 57.53 & 87.77 & 73.20 \\
          w/ Beam Search \cite{sutskever2014sequence} & 87.09 & 74.68 & 39.86 & 55.56 & 98.58 & 61.90 & 92.86 & 55.17 & 81.82 & 63.33 &       & 53.63 & 90.46 & 73.70 \\
          w/ DoLa \cite{DBLP:journals/corr/abs-2309-03883_DOLA} & 86.44 & 73.62 & 43.33 & 61.50 & \textbf{98.87} & 62.86 & \textbf{94.23} & 54.48 & 70.91 & 83.33 &       & 55.42 & 89.20 & 73.82 \\
          w/ VCD \cite{Leng_2024_CVPR_VCD} & 82.24 & 61.81 & 35.28 & 55.31 & 98.58 & 49.52 & 89.42 & 55.17 & 74.55 & 53.33 &       & 48.44 & 83.25 & 67.40 \\
          w/ OPERA \cite{Huang_2024_CVPR_OPERA} & 82.78 & 63.10 & 37.08 & 61.28 & 98.58 & 59.52 & \textbf{94.23} & 60.00 & 74.55 & 53.33 &       & 51.65 & 84.21 & 69.38 \\
    \midrule
    \rowcolor[rgb]{ .851,  .851,  .851} \multicolumn{15}{l}{\textit{Our Proposal}} \\
    CheX-Phi3.5V(20k) & 89.55 & 76.84 & 46.40 & 67.10 & 98.58 & \textbf{65.90} & 91.13 & 80.88 & 66.38 & 73.77 &       & 60.87 & 91.10 & 77.34 \\
    CheX-Phi3.5V(30k) & \textbf{91.39} & \textbf{80.68} & \textbf{49.35} & \textbf{69.12} & \textbf{98.87} & 65.72 & 91.18 & \textbf{90.51} & 66.03 & \textbf{76.54} &       & \textbf{63.12} & \textbf{93.64} & \textbf{79.74} \\
    \bottomrule
    \end{tabular}} 
    \caption{Overall accuracy (\%) on our chest X-ray VQA dataset across question and answer types, compared with general and medical VLMs. \textit{SA} uses short answers; \textit{RA} rewrites rationales via GPT-4o. Our CheX-Phi3.5V uses greedy decoding and \textbf{CheXPO} strategy to generate 20k and 30k preference-aligned data.
    }
  \label{tab:sota-results}
\end{table*}%

\section{Experimental Settings}
\label{sec:exp}

\subsection{Model Variants}
We fine-tune our proposed multi-task chest X-ray VQA dataset using Phi-3.5 Vision, a 4.2B-parameter lightweight vision-language model from the Microsoft Phi-3 series \cite{Abdin2024Phi3TR}. The model integrates a CLIP ViT-L/14 image encoder \cite{Radford2021LearningTV} and adopts the transformer architecture of Phi-3.5 Mini. Our training includes two stages: SFT for instruction following and medical image understanding, and DPO for improving clinical reliability and alignment with response preferences. With LoRA \cite{Hu2021LoRALA} and limited image size configurations (num\_crop $\leq4$), all experiments can be conducted on NVIDIA RTX 4090 24GB GPUs. For the dataset retrieval component, we utilize BioMedCLIP \cite{Zhang2023BiomedCLIPAM}, which employs a ViT-B/16 as the visual encoder and BERT as the text encoder.

\subsection{Experimental Setup and Evaluation}

We apply LoRA tuning (rank 128, alpha 2) and update only the injected layers. Both the SFT and DPO stages use AdamW with a cosine scheduler. For SFT, we set learning rate = 8e-5, batch size = 128, and trained for 2 epochs, which yielded the optimal results; for DPO, we set learning rate = 4e-6, batch size = 256, and \(\beta = 0.1\). For hard example mining, we randomly sample from the training set, stratified by question-answer type pairs, and forward the samples to initialize \(\mathcal{D}_{hard}\). The sampling rate \(\gamma\), log-prob threshold \(\sigma\), and Top-K are adjusted based on the target preference dataset size (see Appendix); for the 30k setting, we set \(\gamma = 2.7\%\), \(\sigma = -0.3\) ( prob \(\approx 0.74\) ), and \(K = 10\). To simplify the evaluation, we assess only the short answer \(\mathbf{y}_{\mathcal{A}}\) in the VLM response. Whether it is a closed-type ``\textit{yes/no}'' answer or an open-type answer, we calculate accuracy using strict string matching.

%% file: sec/5_result.tex

\section{Experimental Results}
\subsection{Comparison with SoTAs}

We compare our method against two categories of vision-language models: 1) \textbf{General-purpose VLMs}, including LLaVA-v1.6 \cite{llava}, Qwen-VL-Chat \cite{qwen}, and Phi-3.5V \cite{Abdin2024Phi3TR}. 2) \textbf{Medical VLMs}, including Med-Flamingo \cite{pmlr-v225-moor23aMed-Flamingo}, RadFM \cite{DBLP:journals/corr/abs-2308-02463_RadFM}, LLaVA-Med \cite{llava-med}, and HuatuoV \cite{chen2024huatuogptvisioninjectingmedicalvisual}. To ensure consistent answer formatting, we provide each model with one-shot exemplars selected according to the question-answer type. We also fine-tune Phi-3.5V on our dataset using two rationale settings (short answer ``SA'' and rewritten rationales ``RA''), and incorporate several decoding or hallucination mitigation strategies, including Beam Search (beam=3) \cite{sutskever2014sequence}, DoLa \cite{DBLP:journals/corr/abs-2309-03883_DOLA}, VCD \cite{Leng_2024_CVPR_VCD}, and OPERA \cite{Huang_2024_CVPR_OPERA}. The results are summarized in Table \ref{tab:sota-results}. Our SFT notably improves performance over the Phi-3.5V baseline (31.3\% $\rightarrow$ 73.2\%), outperforming other models on nearly all question types. Hallucination mitigation strategies are not consistently effective, while the CheXPO strategy brings further gains to the SFT model (73.2\% $\rightarrow$ 79.74\%, 8.93\% relative improvement). Note that these decoding strategies can be jointly applied to CheXPO-enhanced models. Additionally, we provide a comparison of the MIMIC-CXR-VQA \cite{Bae2023EHRXQAAM} and Medical-Diff-VQA \cite{Hu2023ExpertKI_MedDiffVQA} benchmarks in the Appendix, based on classification and $N$-gram evaluation metrics.


\begin{figure}[h]
\centering
\begin{minipage}[b]{0.235\textwidth}
    \centering
    \includegraphics[width=\textwidth]{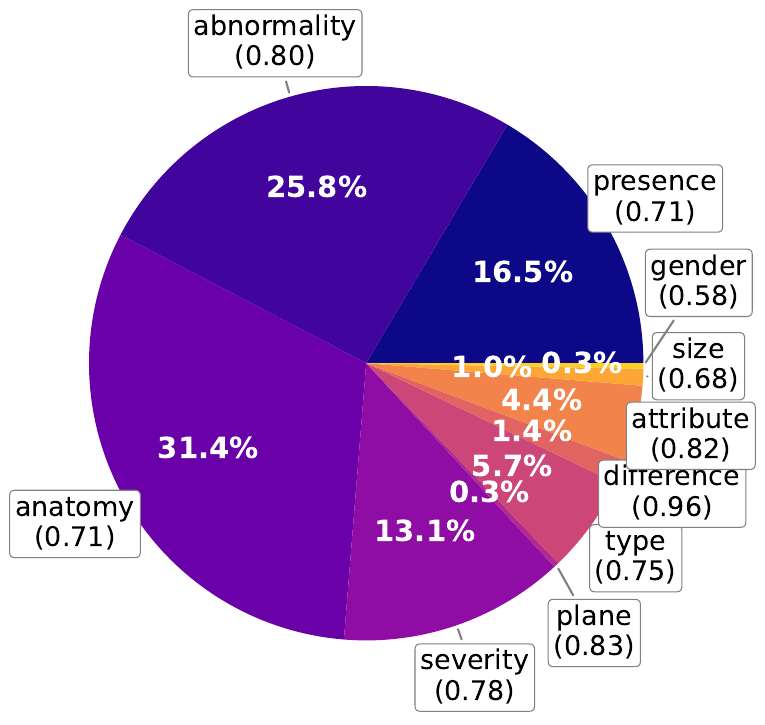}
    \vspace{-0.5cm}
    \caption*{(a)}
\end{minipage}
\hfill
\begin{minipage}[b]{0.235\textwidth}
    \centering
    \includegraphics[width=\textwidth]{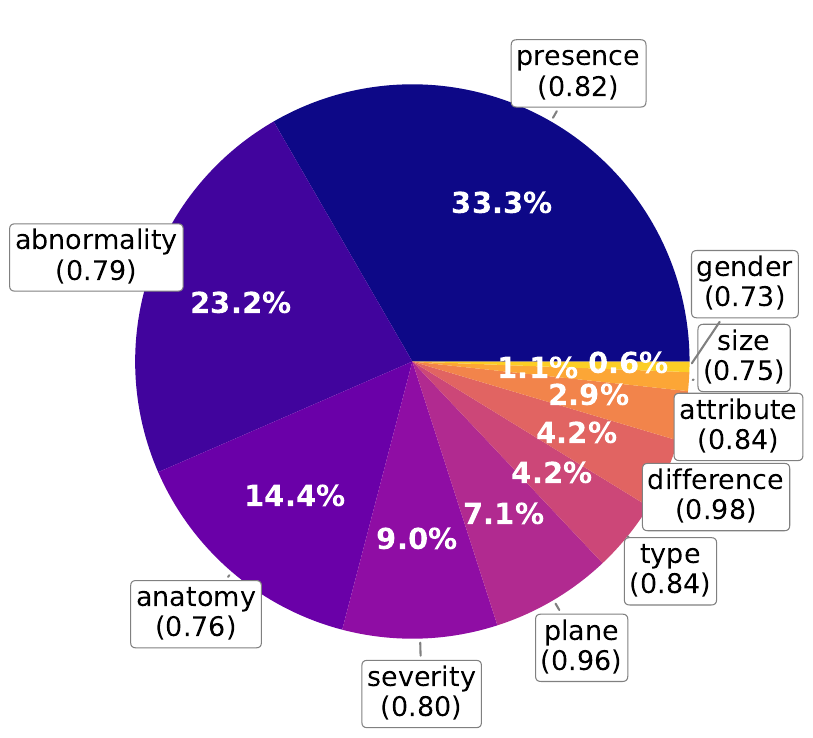}
    \vspace{-0.5cm}
    \caption*{(b)}
\end{minipage}
\vspace{-0.6cm}
\caption[Hard example distribution analysis]{
\textbf{Comparison of distribution between SFT failure cases (a) and the original dataset (b), with the average token-level probability of the answer part.} In the failure cases, abnormality, anatomy, and severity dominate.
}
\label{fig:class_distribution}
\end{figure}

\subsection{Ablation Study}
\label{sec:abl}
To systematically validate the contributions of each component in our confidence-similarity joint mining framework, we conduct extensive ablation studies addressing the following research questions: \textbf{RQ1. }Why do we need hard example mining? \textbf{RQ2. }How does multi-modal retrieval contribute to sample selection? \textbf{RQ3. }Is counterfactual rationale superior to random selection?
\vspace{0.2cm}

\textbf{Necessity of Hard Example Mining.} Figure \ref{fig:class_distribution} (a) and (b) present the distribution of failure case types for the SFT model on the test set, along with the overall distribution across all test data, which combines the three QA subsets: \textbf{Basic}, \textbf{Region}, and \textbf{Comparison}. Additionally, we compute the average token-level probability of the answer \(\mathcal{A}\) in the SFT model responses for each case type. It is observed that the \texttt{Abnormality}, \texttt{Anatomy}, and \texttt{Severity} categories dominate the failure cases, which does not fully align with the original data distribution. Specifically, \texttt{Abnormality} refers to questions about abnormalities, encompassing four categories: anatomical finding, disease, device, and tubes/lines. \texttt{Anatomy} involves open-ended questions requiring anatomical location responses, while \texttt{Severity} pertains to the level of a symptom or disease. Most failure cases exhibit lower probabilities than average, with simple questions such as determining “\textit{whether the view is AP or PA}” (\texttt{Plane} type) being highly reliable. However, some differences between case types are less pronounced. For instance, the probability gap between \texttt{Severity} and \texttt{Attribute} types is only 2\%. More critically, failure cases in the \texttt{Abnormality} category show higher confidence than average, despite being incorrect. This discrepancy underscores the need for hard example mining and data distribution balancing during preference alignment. Without these steps, the model may focus disproportionately on the majority class and continue to generate inaccurate predictions for rare yet critical cases.

\begin{figure}[t]
    \centering
    \includegraphics[width=\columnwidth]{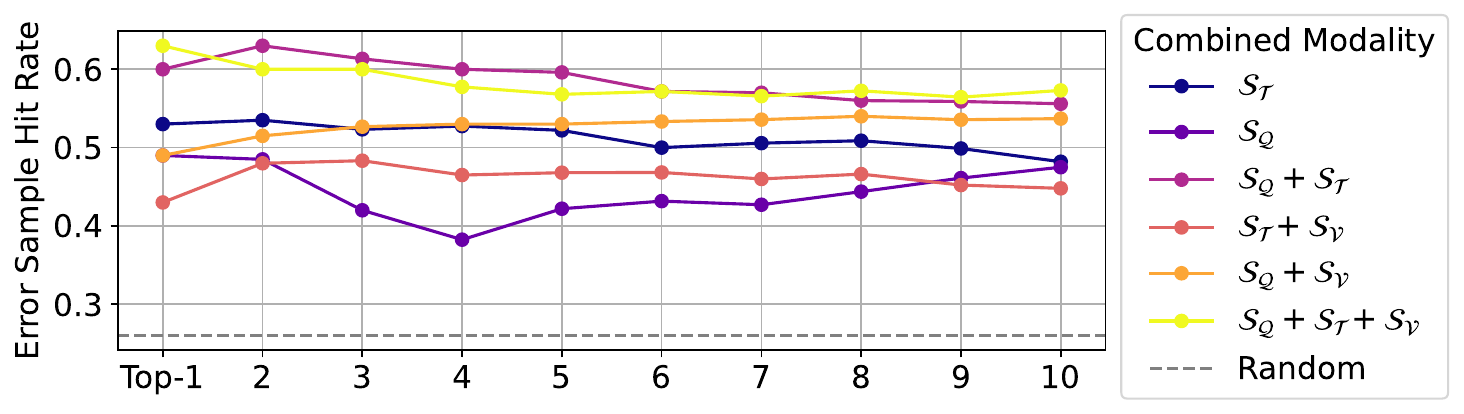}
    \vspace{-0.7cm}
    \caption{\textbf{Retrieval Performance for SFT Failure Cases.} The y-axis represents the proportion of SFT model failure cases in the retrieved top-K samples.}
    \label{fig:retrieval_accuracy}
\end{figure}

\begin{table}[h]
  \centering
  \resizebox{\linewidth}{!}{
    \begin{tabular}{lrrrrr}
    \toprule
    \textbf{Method} & \multicolumn{1}{p{4.895em}}{\textbf{Samples\newline{}Forwarded}} & \multicolumn{1}{p{2.32em}}{\textbf{Hit\newline{}Rate}} & \multicolumn{1}{p{3.785em}}{\textbf{Forward\newline{}Time (s)}} & \multicolumn{1}{p{4em}}{\textbf{Retrieval\newline{}Time (s)}} & \multicolumn{1}{p{3.68em}}{\textbf{Total\newline{}Time (s)}} \\
    \midrule
    \textbf{Direct SFT Forward} & 4k    & $\sim$25\% & 6000  & N/A   & 6000 \\
    \textbf{Retrieval + Forward} & 1.8k  & $\sim$55\% & 2700  & 30    & 2730 \\
    \bottomrule
    \end{tabular}%
  }
  
  \caption{Efficiency Comparison of hard example mining. SFT model inference takes 1.5s per sample, while retrieval time is almost negligible (5$\sim$15ms per sample).  }
  \label{tab:efficiency}
\end{table}

\vspace{-0.5cm}

\textbf{Effect of Multi-Modal Retrieval.}
The retrieval-based approach for hard example mining is employed primarily for two key reasons: efficiency and maintaining diversity while ensuring semantic relevance. Therefore, we evaluate the effectiveness of multimodal retrieval from these two aspects. Figure~\ref{fig:retrieval_accuracy} presents the proportion of hard examples retrieved using different methods from Top-1 to Top-10, including retrieval based on question-only \(\mathcal{S}_\mathcal{Q}\), rationale-only \(\mathcal{S}_\mathcal{T}\), and their combinations with image features \(\mathcal{S}_\mathcal{V}\). The query set consists of 100 SFT failure cases sampled from the training set with a balanced QA-type distribution, while the gallery set comprises 20k samples uniformly drawn from the training set, also with a balanced QA-type distribution. The results align well with intuition: queries that share similar semantics in questions, rationales, or visual characteristics with existing SFT failure cases are more likely to be mispredicted by the SFT model. Given that retrieval-based hard example mining achieves a significantly higher hit rate than random sampling, it greatly improves the efficiency of data selection, as shown in Table~\ref{tab:efficiency}.

\begin{figure}[h]  
\centering  
\includegraphics[width=1\columnwidth]{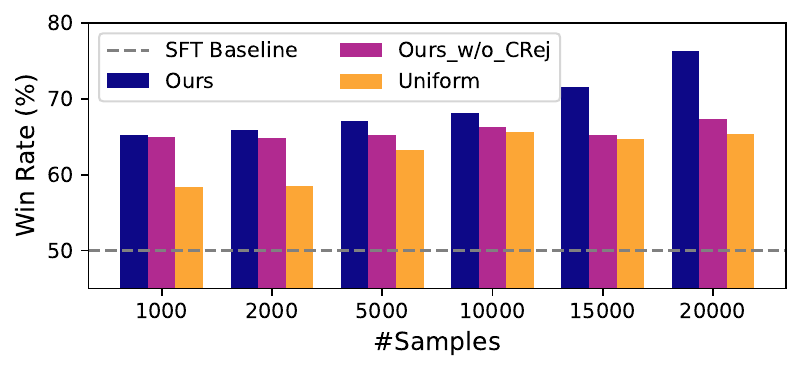}  
\vspace{-0.7cm}
\caption[Data efficiency comparison]{  
\textbf{Win rates relative to the SFT baseline across different sampling methods and preference set sizes. Correct predictions with \(\sigma > -0.3\) are excluded from Top-10 retrieval.
} }  
\label{fig:data_scaling}  
\end{figure}

\begin{figure}[h]  
\centering  
\includegraphics[width=1\columnwidth]{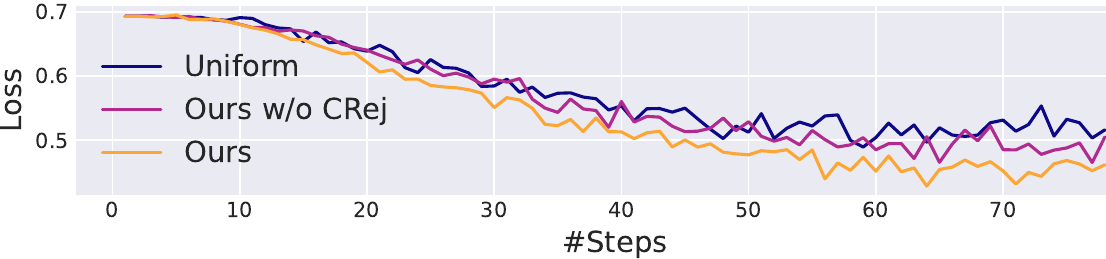}  
\vspace{-0.7cm}
\caption[Data efficiency comparison]{  
\textbf{Training loss across three sampling strategies (20k samples, batch size=256).}}  
\label{fig:loss}  
\end{figure}

\begin{figure}[h]  
\centering  
\includegraphics[width=0.9\columnwidth]{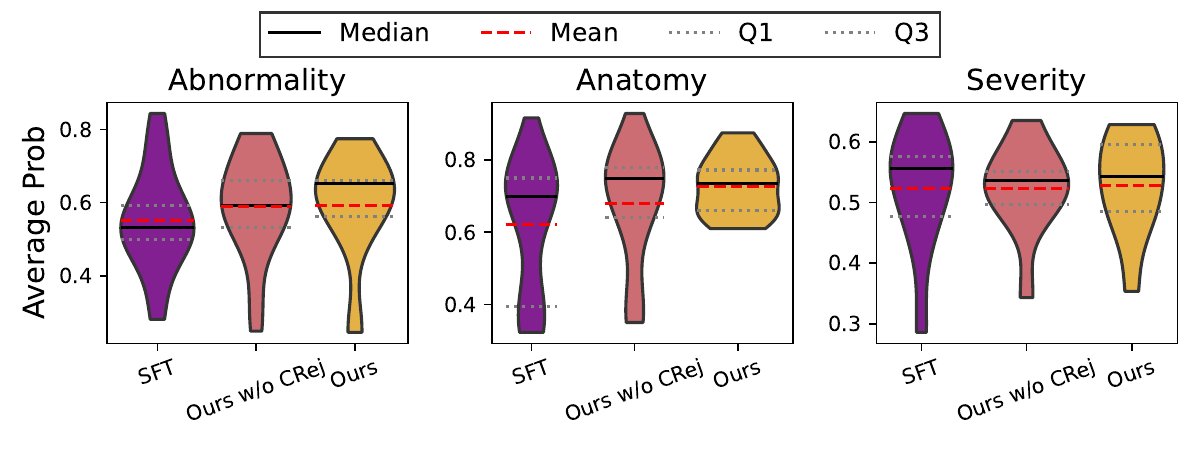}  
\vspace{-0.4cm}
\caption{
Distribution of answer-averaged probabilities for revised responses on three representative question types (Abnormality, Anatomy, Severity), where the SFT model failed but the proposed model succeeded. 
}
\label{fig:rationality_scores}  
\end{figure}

\textbf{Effect of Counterfactual Rationale.} Although directly retrieving failure cases helps identify clear weaknesses of the model, a substantial portion of low-confidence correct predictions still reflect uncertain or fragile decision boundaries. These cases may not be outright errors but indicate areas where the model lacks robustness or consistency, and thus require further preference optimization. To address this, we apply preference alignment with different data selection strategies, as shown in Figure \ref{fig:data_scaling}, \ref{fig:loss} and \ref{fig:rationality_scores}. Figure \ref{fig:data_scaling} shows that although \textbf{\textcolor[HTML]{E3A939}{uniformly sampled preference data}} brings initial gains over the SFT baseline with just 1k examples, its effectiveness plateaus with more data. In contrast, our \textbf{\textcolor[HTML]{B05BA2}{hard example mining strategy}} consistently improves performance by focusing on informative failure cases, but also begins to saturate beyond 10k samples due to limited diversity in failure types. Our \textbf{\textcolor[HTML]{1F3B8D}{full strategy}}—hard example mining with counterfactual rationale—further improves win rates across all data sizes, particularly beyond 10k, demonstrating its ability to break through the plateau by synthesizing semantically coherent but clinically incorrect negative samples. Figure \ref{fig:loss} confirms this advantage from a training perspective: our method yields lower training loss, suggesting more effective preference learning. Furthermore, Figure \ref{fig:rationality_scores} provides a token-level analysis on three representative question types. Compared to SFT, our method increases the model’s confidence for previously uncertain cases (e.g., \texttt{Anatomy}, \texttt{Severity}), while also correcting confidently wrong predictions in Abnormality-type questions. This highlights the importance of counterfactual rationale in improving both reliability and calibration.

\vspace{-0.2cm}

\vspace{-0.2cm}

\begin{figure*}[t]
    \centering
    \includegraphics[width=0.95\textwidth]{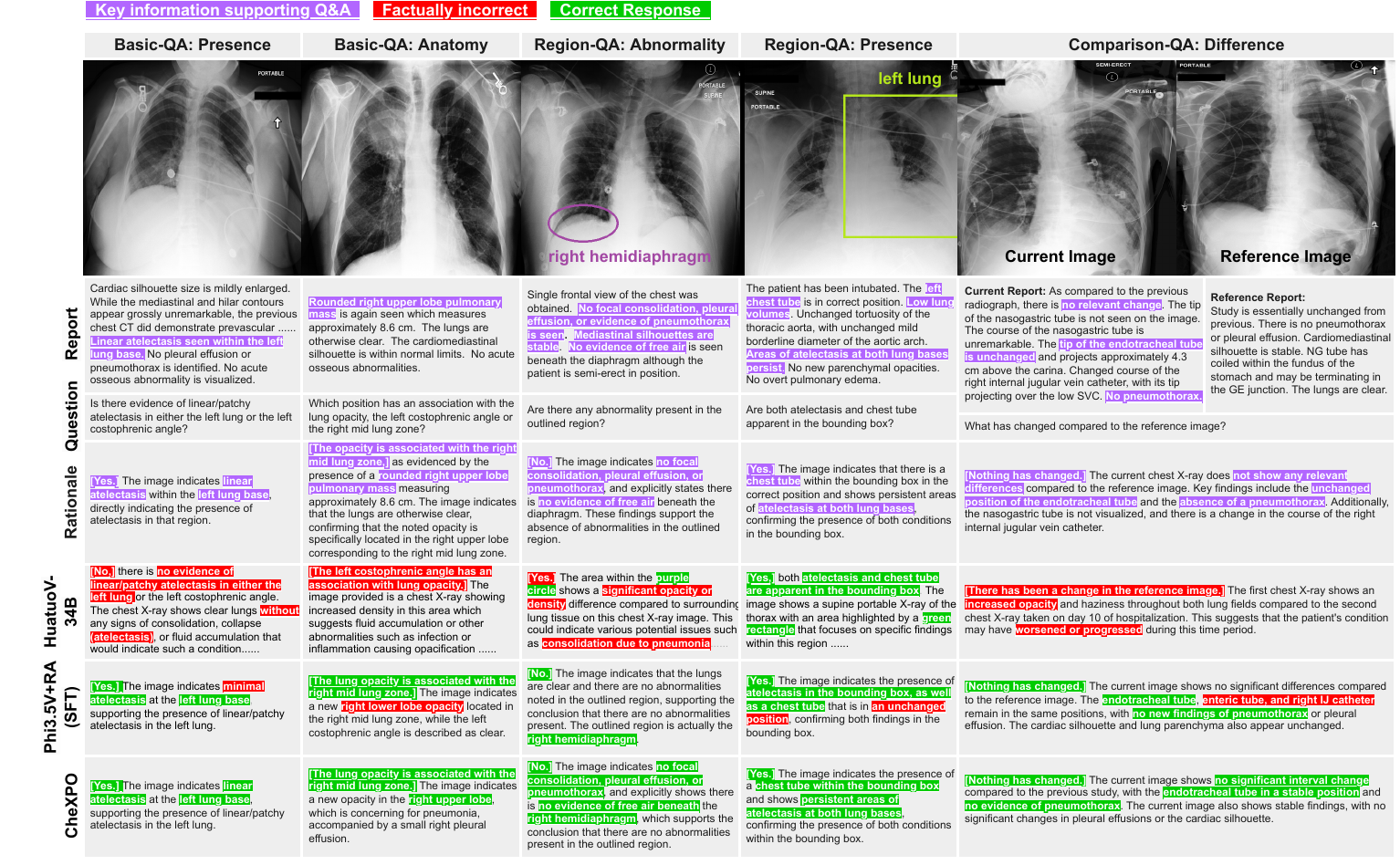}
    \vspace{-0.3cm}
    \caption{Representative chest X-ray cases across three major tasks (Basic-QA, Region-QA, Comparison-QA), comparing outputs from HuatuoV-34B, our SFT, and CheXPO models. }
    \label{fig:case_study}
\end{figure*}


\vspace{-0.5cm}

\subsection{Case Study}

\setlength{\fboxsep}{1pt} 
Figure \ref{fig:case_study} presents representative cases from three major chest X-ray VQA tasks: Basic-QA, Region-QA, and Comparison-QA. Model outputs include HuatuoV-34B, currently the best-performing MedVLM, alongside our fine-tuned SFT and CheXPO models. Region annotations (\textbf{\textcolor[HTML]{A349A4}{circles}} and \textbf{\textcolor[HTML]{B5E61D}{rectangles}}) correspond to the Chest ImaGenome \cite{Wu2021ChestID_ChestImaGenome} silver standard labels. The \colorbox[HTML]{00CC00}{\textcolor{white}{\textbf{analysis}}} highlights that our proposed CheXPO training effectively reinforces detailed medical reasoning within the generated rationales. Specifically, CheXPO enhances the precision of anatomical descriptions and pathological details, producing explanations that closely align with clinical observations, as exemplified by accurately identifying anatomical landmarks and distinguishing pathological nuances. In comparison, HuatuoV-34B exhibits severe hallucinations in its generated explanations. Notably, HuatuoV-34B demonstrates proficiency in recognizing visual prompts directly overlaid on the images. This ability may stem from its pre-trained visual encoder or exposure to physician-annotated prompts during PubMed-based training. We also observe that although the SFT model occasionally arrives at the correct answer, its reasoning process is often flawed or inconsistent, raising concerns about the validity of current evaluation methods for medical VLMs. This mirrors challenges faced in hallucination evaluation for VLMs, where explanation quality is difficult to assess—highlighting the urgent need for more robust and clinically grounded evaluation benchmarks.

%% file: sec/6_conclu.tex
\section{Conclusion}
In this paper, we present a comprehensive pipeline—from dataset construction to preference optimization—for building reliable and interpretable chest X-ray VLMs. We unify multiple benchmarks into a multi-task VQA dataset covering basic, region, and comparison tasks, paired with expert-style rationales. To address critical failure patterns such as hallucinated findings and anatomical misjudgments, we propose a scalable confidence-similarity joint mining strategy and a counterfactual rationale reinforcement approach, named CheXPO, for preference alignment. Our model outperforms existing general and medical VLMs across all tasks, achieving significant gains in clinical reliability and interpretability without relying on additional expert annotations. We believe this framework provides a strong foundation for the future development of safe and trustworthy medical AI systems.

\section*{Acknowledgement}

This work was supported in part by the National Science and Technology Major Project under Grant 2022ZD0117103, in part by the Outstanding Youth Science Foundation of Shaanxi Province under Grant 2025JC-JCQN-083, in part by the National Natural Science Foundation of China under Grants 62172222 and 62302354, in part by the Key Research and Development Program of Shaanxi Province under Grants 2025CY-YBXM-047 and 2024GX-YBXM-039, and in part by the Fundamental Research Funds for the Central Universities under Grant QTZX23084.

%% file: sec/7_appendix.tex
\section{Data Sources}
We integrate data from four sources: MIMIC-CXR \cite{Johnson2019_MIMICCXRAD} for images and reports, Chest ImaGenome \cite{Wu2021ChestID_ChestImaGenome} for labeled organs and diseases, and both MIMIC-CXR-VQA \cite{Bae2023EHRXQAAM} and Medical-Diff-VQA \cite{Hu2023ExpertKI_MedDiffVQA} for QA pairs involving single and comparative images. Below are brief descriptions of each dataset:

\textbf{MIMIC-CXR (v2.0.0)} \cite{Johnson2019_MIMICCXRAD} is a large-scale, publicly available dataset containing 377,110 chest radiographs with structured labels derived from 227,827 free-text radiology reports, collected at BIDMC between 2011 and 2016.

\textbf{Chest ImaGenome (v1.0.0)} \cite{Wu2021ChestID_ChestImaGenome} provides scene graph annotations for 242,072 frontal chest X-rays from MIMIC-CXR, explicitly representing anatomical locations and their associated attributes. It comprises two subsets: the automatically generated silver dataset covering all images, and the clinician-validated gold dataset derived from 500 unique patients.

\textbf{MIMIC-CXR-VQA} \cite{Bae2023EHRXQAAM} is a medical visual question answering dataset containing approximately 377,000 QA pairs associated with chest X-rays. This dataset was constructed by integrating MIMIC-CXR, MIMIC-IV \cite{Johnson2023MIMICIVAF}, and Chest ImaGenome. It includes 48 unique question templates categorized into seven clinical domains: existence, anatomy, attribute, abnormality, size, plane, and gender. Each template provides approximately 16.5 paraphrases generated by GPT-4 under clinical expert guidance.

\textbf{Medical-Diff-VQA} \cite{Hu2023ExpertKI_MedDiffVQA} is a medical difference visual question answering dataset, constructed from MIMIC-CXR. It contains 700,703 QA pairs from 164,324 image pairs, organized into seven clinically relevant categories: abnormality, location, type, level, view, presence, and difference. It involves chest radiograph comparisons to support comparative clinical reasoning aligned with standard medical practices.

All the above datasets are publicly available on PhysioNet platform \cite{PhysioNet}, requiring credentialed access approval.

\section{Data Statistics}
This section provides a detailed statistical overview of the datasets used for training and evaluation. The key aspects of the dataset statistics include the number of samples for different subtasks, the distribution of training, validation, and test sets, and the ratio of closed and open-ended questions in each dataset. Table \ref{tab:dataset} outlines the key dataset statistics.

\begin{figure}[htbp]
  \centering
  \includegraphics[width=0.7\columnwidth]{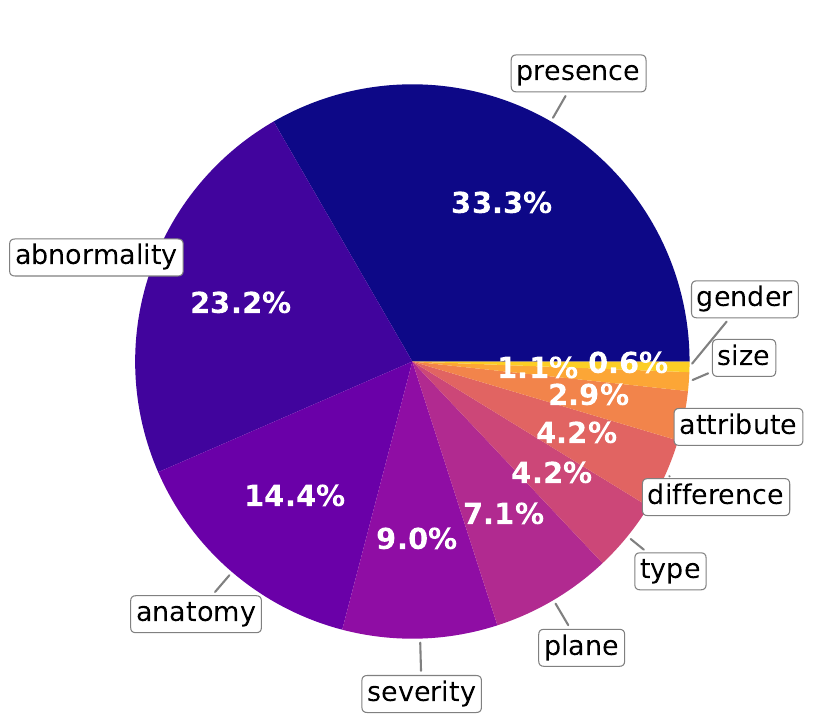}
  \caption{Distribution of question types in the dataset.}
  \label{fig:pie_chart}
\end{figure}

The MIMIC-CXR-VQA\cite{Bae2023EHRXQAAM} dataset provides seven distinct question types: \textit{Presence}, \textit{Anatomy}, \textit{Attribute}, \textit{Abnormality}, \textit{Size}, \textit{Plane}, and \textit{Gender}. The Medical-Diff-VQA \cite{Hu2023ExpertKI_MedDiffVQA} dataset, while also offering seven question types, has some differences compared to MIMIC-CXR-VQA. The question types in this dataset include \textit{Abnormality}, \textit{Presence}, \textit{View}, \textit{Location}, \textit{Severity}, \textit{Type}, and \textit{Difference}. Given the overlap between some categories, such as \textit{Location} and \textit{Anatomy}, and \textit{View} and \textit{Plane}, these types have been merged in the final classification, resulting in ten distinct categories overall, as shown in Figure~\ref{fig:pie_chart}. The aligned question types and their corresponding examples are summarized in Table \ref{tab:question_types_examples}.

\begin{table}[t]
  \centering
  \resizebox{\columnwidth}{!}{
    \begin{tabular}{l|l|l|rrrr}
    \toprule
    \textbf{Subtask} & \multicolumn{1}{l|}{\textbf{Anno. Source}} & \textbf{Split} & \multicolumn{1}{l}{\textbf{QA Pairs}} & \multicolumn{1}{l}{\textbf{Closed}} & \multicolumn{1}{l}{\textbf{Open}} & \multicolumn{1}{l}{\textbf{Images}} \\
    \midrule
    \multirow{3}[2]{*}{\textbf{Basic}} & \multicolumn{1}{p{8.465em}|}{MIMIC-CXR-VQA} & Train & 181,349 & 117,394 & 63,955 & 91,406 \\
          & \multicolumn{1}{l|}{MIMIC-CXR} & Valid & 44,264 & 28,597 & 15,667 & 6,201 \\
          &       & Test  & 9,066 & 5,704 & 3,362 & 409 \\
    \midrule
    \multirow{3}[2]{*}{\textbf{Region}} & \multicolumn{1}{l|}{MIMIC-CXR-VQA} & Train & 44,686 & 34,064 & 10,622 & 36,293 \\
          & \multicolumn{1}{l|}{MIMIC-CXR} & Valid & 10,574 & 8,061 & 2,513 & 5,009 \\
          & \multicolumn{1}{l|}{Chest ImaGenome} & Test  & 2,134 & 1,605 & 529   & 408 \\
    \midrule
    \multirow{3}[2]{*}{\textbf{Compare}} & \multicolumn{1}{l|}{Medical-Diff-VQA} & Train & 319,859 & 165,089 & 154,770 & 110,761 \\
          & \multicolumn{1}{l|}{MIMIC-CXR} & Valid & 14,176 & 7,202 & 6,974 & 5,827 \\
          &       & Test  & 13,646 & 6,901 & 6,745 & 5,706 \\
    \bottomrule
    \end{tabular}%
    }

  \caption{Dataset statistics. "Closed" refers to yes/no questions, while "Open" includes other answer types. }
  \label{tab:dataset}%
\end{table}%

\begin{table}[!t]
  \centering
  \resizebox{\columnwidth}{!}{
    \begin{tabular}{lp{25em}}
    \toprule
    \textbf{Question Type} & \textbf{Example} \\
    \midrule
    Presence & Is any technical assessments present within the left breast? \\
    Anatomy & Enumerate all the regions in the anatomy that are related to vascular congestion. \\
    Attribute & Provide a list of all the tubes/lines found in both the carina and neck. \\
    Abnormality & Can you identify any abnormalities? \\
    Size  & Is the cardiac silhouette wider than 50\% of the thorax's width? \\
    Plane & Does this image have an AP projection? \\
    Gender & Is the patient identified as male? \\
    Severity & What level is the pleural effusion? \\
    Type  & What type is the edema? \\
    Difference & What has changed compared to the reference image? \\
    \bottomrule
    \end{tabular}%
  }
  \caption{Question Types and Examples}
  \label{tab:question_types_examples}
\end{table}

\begin{table*}[!t]
  \centering
  \resizebox{\linewidth}{!}{%
    \begin{tabular}{llllllllcllcc}
    \toprule
    \multirow{2}[4]{*}{\textbf{Experiment }} &       & \multicolumn{4}{c}{\textbf{Optimization}} &       & \multicolumn{2}{c}{\textbf{DPO}} &       & \multicolumn{3}{c}{\textbf{LoRA}} \\
\cmidrule{3-6}\cmidrule{8-9}\cmidrule{11-13}          &       & \textbf{Learning Rate} & \textbf{Batch Size} & \textbf{Epoch} & \textbf{Optimizer} &       & \textbf{Beta} & \multicolumn{1}{l}{\textbf{Loss\_type}} &       & \textbf{Rank} & \multicolumn{1}{l}{\textbf{Alpha}} & \multicolumn{1}{p{8.965em}}{\textbf{Target modules}} \\
\cmidrule{1-1}\cmidrule{3-6}\cmidrule{8-9}\cmidrule{11-13}    SFT   &       & 6e-5, \textbf{8e-5} & 128   & \textbf{1}, 2  & AdamW &       & ——    & \multicolumn{1}{l}{——} &       & 64, \textbf{128}, 256 & \multirow{3}[2]{*}{2} & q\_proj, k\_proj, v\_proj, \\
    DPO (20k) &       & 3e-6, \textbf{4e-6}, 5e-6, 8e-6 & 256   & 1     & AdamW &       & \textbf{0.1}, 0.15,0.2,0.3 & \multirow{2}[1]{*}{ipo, \textbf{sigmoid}, hinge, robust} &       & \textbf{128}, 256 &       & o\_proj, gate\_proj, \\
    DPO (30k) &       & 3e-6, 4e-6, \textbf{5e-6}, 8e-6 & 256   & 1     & AdamW &       & 0.1, \textbf{0.15},0.2,0.3 &       &       & \textbf{128}, 256 &       & up\_proj, down proj \\
    \bottomrule
    \end{tabular}%
  }
  \caption{Optimization Parameters for SFT and DPO Experiments with LoRA} 
  \label{tab:parameter}
\end{table*}

\section{Training Hyperparameters and Model Setup}

This section outlines the hyperparameters used during the training process, including learning rates, batch sizes, and the optimizer settings. Additionally, we provide details on the architecture of the models, including the choice of neural network and its specific configurations. The training setup is designed to balance computational efficiency and model performance. Table \ref{tab:parameter} shows the optimization parameters used for SFT and DPO experiments with LoRA. The optimal parameter settings for each training phase are highlighted in \textbf{bold}. In the SFT phase, increasing the number of epochs does not continuously improve performance. Additionally, modifying the LoRA rank (either increasing or decreasing) results in suboptimal performance. For the DPO phases, various settings for learning rate and beta values were tested, with the best configurations indicated.

\begin{table}[H]
  \centering
    \begin{tabular}{lrr}
    \toprule
    \textbf{Model} & \multicolumn{2}{c}{\textbf{Test set }} \\
\cmidrule{2-3}          & \multicolumn{1}{l}{Acc} & \multicolumn{1}{l}{F1 (micro)} \\
    \midrule
    Prior (Most) & 25.4  & 0.25 \\
    Prior (Question) & 32.4  & 0.32 \\
    PubMedCLIP \cite{Eslami2023PubMedCLIPHM} & 54.9  & 0.54 \\
    PubMedCLIP* \cite{Eslami2023PubMedCLIPHM} & 56.5  & 0.56 \\
    MedViLL* \cite{Moon2021MultiModalUAMedViLL} & 63.6  & 0.67 \\
    M3AE \cite{Chen2022MultimodalMAM3AE}  & 68.0  & 0.68 \\
    M3AE* \cite{Chen2022MultimodalMAM3AE} & 69.2  & 0.73 \\
    \midrule
    Phi3.5V+SA & 67.4  & 0.70 \\
    Phi3.5V+RA & 71.5  & 0.74 \\
    CheX-Phi3.5V(20k) & 75.2  & 0.78 \\
    CheX-Phi3.5V(30k) & 81.8  & 0.84 \\
    \bottomrule
    \end{tabular}%
\caption{Comparison of various models on the MIMIC-CXR-VQA benchmark. The results include accuracy (Acc) and micro-averaged F1 scores for each model. *Note: This is a re-implemented result from \cite {Bae2023EHRXQAAM}.}
\label{tab:classify_sota}
\end{table}%

\begin{table*}[!h]
  \centering
    \begin{tabular}{lrrrrrrr}
    \toprule
    \textbf{Methods} & \multicolumn{1}{l}{\textbf{Bleu-1}} & \multicolumn{1}{l}{\textbf{Bleu-2}} & \multicolumn{1}{l}{\textbf{Bleu-3}} & \multicolumn{1}{l}{\textbf{Bleu-4}} & \multicolumn{1}{l}{\textbf{METEOR}} & \multicolumn{1}{l}{\textbf{ROUGE-L}} & \multicolumn{1}{l}{\textbf{CIDEr}} \\
    \midrule
    MCCFormers* \cite{Qiu2021DescribingALMCCFormer} & 0.214 & 0.190 & 0.170 & 0.153 & 0.319 & 0.340 & 0.000 \\
    IDCPCL* \cite{Yao2022ImageDCIDCPCL} & 0.614 & 0.541 & 0.474 & 0.414 & 0.303 & 0.582 & 0.703 \\
    EKAID \cite{Hu2023ExpertKI_MedDiffVQA} & 0.628 & 0.553 & 0.491 & 0.434 & 0.339 & 0.577 & 1.027 \\
    \midrule
    Phi3.5V+SA & 0.673 & 0.618 & 0.516 & 0.503 & 0.413 & 0.711 & 1.927 \\
    Phi3.5V+RA & 0.685 & 0.629 & 0.525 & 0.513 & 0.417 & 0.723 & 1.953 \\
    CheX-Phi3.5V(20k) & 0.700 & 0.648 & 0.549 & 0.539 & 0.421 & 0.735 & 2.001 \\
    CheX-Phi3.5V(30k) & 0.719 & 0.679 & 0.575 & 0.564 & 0.426 & 0.752 & 2.080 \\
    \bottomrule
    \end{tabular}%
  \caption{Comparison of various methods on text generation evaluation metrics of Medical-Diff-VQA. *Note: This is a re-implemented result from \cite {Hu2023ExpertKI_MedDiffVQA}.}
  \label{tab:bleu_sota}%
\end{table*}%

\section{Comparison with Existing Benchmarks}
We compare the performance of our model with the existing benchmarks of MIMIC-CXR-VQA \cite{Bae2023EHRXQAAM} and Medical-Diff-VQA \cite{Hu2023ExpertKI_MedDiffVQA}. Table \ref{tab:classify_sota} compares several existing classification methods on the MIMIC-CXR-VQA benchmark, including models like PubMedCLIP, MedViLL, and others. Our model is evaluated using only the short answer portion $\mathbf{y}_\mathcal{A}$ of the model response (e.g. ``\textit{Yes}'' or ``\textit{Cardiac silhouette and the mediastinum}''). It is important to note that the dataset contains some questions with multiple answers (e.g. \texttt{[``cardiac silhouette'', ``mediastinum'']}). The accuracy metric is calculated by checking if the predicted short answer strictly matches the ground truth answer. The micro-averaged F1 score is calculated by considering the total true positives, false positives, and false negatives across all instances, providing a balanced evaluation of both precision and recall. Table \ref{tab:bleu_sota} compares the performance of several methods, including MCCFormers, IDCPCL, EKAID, and our methods, on standard text generation evaluation metrics for the Medical-Diff-VQA dataset. The methods were evaluated using BLEU scores (ranging from BLEU-1 to BLEU-4), METEOR, ROUGE-L, and CIDEr. As in Table \ref{tab:classify_sota}, only the predicted short answer \(\mathbf{y}_{\mathcal{A}}\) was used for evaluation. The BLEU scores focus on n-gram overlap, METEOR measures semantic similarity, ROUGE-L evaluates longest common subsequences, and CIDEr assesses the consensus between generated and reference text. These metrics were chosen because some answers in the Medical-Diff-VQA dataset are long sentences (e.g., “\textit{The main image is missing the finding of atelectasis compared to the reference image.}”).

\begin{table}[H]
  \centering
    \begin{tabular}{rrrrrr}
        \toprule
        \textbf{$\gamma$} & \textbf{TopK} & \textbf{$\sigma$} & \textbf{\# SFT Fails} & \textbf{\#CRej.} & \textbf{\#Total} \\
        \midrule
        0.9\% & 1     & -0.3  & 0.9k  & 0.1k  & 1k \\
        0.9\% & 2     & -0.3  & 1.7k  & 0.3k  & 2k \\
        0.9\% & 5     & -0.3  & 4.2k  & 0.8k  & 5k \\
        0.9\% & 10    & -0.3  & 8.5k  & 1.5k  & 10k \\
        1.4\% & 10    & -0.3  & 13k   & 2k    & 15k \\
        1.8\% & 10    & -0.3  & 16k   & 4k    & 20k \\
        2.7\% & 10    & -0.3  & 25k   & 5k    & 30k \\
        \bottomrule
    \end{tabular}%
    \caption{Sampling parameters for preference-aligned data with different sample sizes.}
    \label{tab:data_sample}
\end{table}

\section{Data Sampling and Size Considerations}
In this section, we discuss the sampling parameters and sample proportions for different preference-aligned data used in the experiments, as shown in Table \ref{tab:data_sample}. The \(\gamma\) represents the uniform random sampling ratio from the original training set \(\mathcal{D}\), while TopK refers to the multi-modal retrieval described in Equation (4). The \(\sigma\) is the threshold for the log-prob $\mathbf{p}$ in Equation (1), where only the correct samples with a log-prob $\mathbf{p}$ lower than \(\sigma\) are used to generate contrastive rejection \(\mathbf{y}_l\). The "\textbf{\#SFT Fails}" column represents the use of incorrect predictions from the SFT model as \(\mathbf{y}_l\), while "\textbf{\#CRej}" indicates the use of the method described in Subsection 4.3 for generating \(\mathbf{y}_l\). Data samples ranging from 1k to 20k were used in ablation experiments, and the 30k sample size achieved the best results.

\section{Data Construction and Preprocessing}
The data construction process is divided into two steps. The first step involves generating rationales for different subtasks: \textbf{Basic-QA}, \textbf{Region-QA}, and \textbf{Compare-QA}. For the Basic-QA and Region-QA subtasks, rationales are generated using GPT-4o by extracting relevant information from the report to answer the question and provide a short answer, followed by an explanation. The rationale is constructed from the perspective of a doctor examining the images, with the short answer leading the explanation. For the Compare-QA subtask, two images are used: the main image corresponds to the current query, while the reference image is used for comparison. Rationales for this task are extracted from both reports.

The second step involves generating region-based prompts. Using a template containing the placeholder \texttt{\$\{object\}}, the bounding box of the relevant object is located in the Chest Imagenome \cite{Wu2021ChestID_ChestImaGenome} dataset by matching the corresponding \textit{study\_id}. After locating the bounding box, the region of interest (ROI) and relevant words like “\texttt{specified region}” are used to modify the prompt. Additionally, visual prompts are randomly added to the bbox corresponding to the image. The prompts for the basic/region and comparison subtasks are constructed as shown in Figure \ref{lst:prompt_basic_region} and Figure \ref{lst:prompt_compare}, and the attributes included in each data sample are shown in Figure \ref{lst:data_attributes_example}.

Additionally, we emphasize that all rationales in CheXPO are not generated by GPT-4 alone; rather, they are strictly derived from the original radiology reports written by clinical experts in the MIMIC-CXR dataset. To further validate their reliability, we uniformly sampled 1,000 generated counterfactual rationales. These underwent automated assessment, including Disease Correctness (measuring entity overlap with reports), Semantic Similarity (assessing rephrased information relevance), and LLM Evaluation (factual correctness scoring by GPT-4o). A subset of 100 rationales was additionally subjected to human evaluation by medical professionals for factual correctness.

\FloatBarrier 
\begin{table*}[t]
  \centering
  \small
  \renewcommand{\arraystretch}{1.1}   
  \setlength{\tabcolsep}{6pt}        
  \begin{tabular}{p{0.23\linewidth} p{0.43\linewidth} p{0.16\linewidth} p{0.10\linewidth}}
    \toprule
    \textbf{Method} & \textbf{Description} & \textbf{Metric} & \textbf{Result} \\
    \midrule
    Disease Correctness (14-Classes) &
      Extract disease entities (e.g., ``\textit{Atelectasis}'') from rationale and report. Note: Prioritizes generated rationale's correctness over full report coverage. &
      Macro-Avg Precision & 0.8405 \\
    & & Micro-Avg Precision & 0.8868 \\
    \midrule
    Semantic Similarity &
      Encode ``question + rationale'' and full report with Qwen-3-Embedding-0.6B (rationales heavily re-phrased). &
      Cosine Similarity & 0.6844 \\
    \midrule
    LLM Evaluation &
      GPT-4o evaluates ``question + rationale'' against the report (1 = completely incorrect, 5 = completely correct). &
      Score 1–5 & 4.16 \\
    \midrule
    Human Evaluation (n = 100) &
      Radiologists judge factual correctness of each rationale. &
      Score 1–5 & 4.65 \\
    \bottomrule
  \end{tabular}
  \caption{Quality evaluation of 1 000 sampled counterfactual rationales (100 clinician-checked). Higher is better.}
  \label{tab:rationale_quality}
\end{table*}

\section{Preference Optimisation Ablations}
We also experimented with several other preference optimization (PO) methods, including SimPO (Simple Preference Optimization), CPO (Contrastive Preference Optimization), ORPO (Odds Ratio Preference Optimization), and IPO (Identity Preference Optimization). All of these methods were implemented using the TRL library. As we understand, DPO's core is directly relating decision and reward functions within RLHF, avoiding explicit reward modeling. Subsequent PO advancements primarily focus on memory efficiency (e.g., no reference model) or optimizing objectives for enhanced stability and performance. While hyperparameter search wasn't exhaustive, all tested PO methods consistently improved over SFT, and the proposed hard example mining and counterfactual rationale also demonstrated effectiveness. The results are summarised in Table \ref{tab:po_methods}. Even without an extensive hyper-parameter sweep, SimPO and CPO—two alternative PO objectives—achieved accuracy comparable to DPO. Importantly, every evaluated PO variant exhibited a marked performance gain over the SFT baseline, underscoring the robustness of preference-optimised training for chest-X-ray VLMs.

\begin{table*}[!t]
  \centering
  \small
  \renewcommand{\arraystretch}{1.1}
  \setlength{\tabcolsep}{6pt}
  \begin{tabular}{p{0.13\linewidth}p{0.11\linewidth}p{0.10\linewidth}p{0.12\linewidth}p{0.46\linewidth}}
    \toprule
    \textbf{PO Method} & \textbf{Data Size} & \textbf{Overall Acc.\ (\%)} & \textbf{Win Rate\footnotemark{} (\%)} & \textbf{Description / Key Parameters} \\
    \midrule
    SFT  & 544k & 73.2 & —   & Base model trained on our chest-X-ray visual-instruction dataset \\
    \midrule
    DPO (Uniform)       & 20k pairs & 75.5 & 65.4 & Uniform sampling, no hard-example mining \\
    DPO (w/o CRej)      & 20k pairs & 75.7 & 67.3 & Hard-example mining \\
    DPO (Ours)          & 20k pairs & 77.3 & 76.2 & Hard-example mining $+$ counterfactual rationale \\
    \midrule
    SimPO (Uniform)     & 20k pairs & 74.8 & 62.4 & Sensitive to noise \\
    SimPO (w/o CRej)    & 20k pairs & 75.9 & 63.4 & Better for ``chosen $>$ rejected'' pairs \\
    SimPO (Ours)        & 20k pairs & 76.4 & 65.2 & $\text{lr}=4\!\times\!10^{-6}$, $\beta=10$, $\gamma=5$ \\
    \midrule
    CPO  (Uniform)      & 20k pairs & 75.6 & 67.3 & Optimises reference model; increases chosen-sample weight \\
    CPO  (w/o CRej)     & 20k pairs & 76.5 & 69.9 & $\text{lr}=4\!\times\!10^{-6}$, $\beta=0.1$, $\lambda=1.0$ \\
    CPO  (Ours)         & 20k pairs & 77.1 & 72.6 &  \\
    \midrule
    ORPO (Uniform)      & 20k pairs & 75.8 & 70.1 & One-shot SFT $+$ PO \\
    ORPO (w/o CRej)     & 20k pairs & 76.2 & 68.9 & $\text{lr}=1\!\times\!10^{-5}$, $\lambda=0.5$ \\
    ORPO (Ours)         & 20k pairs & 76.8 & 71.4 &  \\
    \midrule
    IPO  (Uniform)      & 20k pairs & 75.4 & 68.2 & Replaces DPO BCE with MSE loss \\
    IPO  (w/o CRej)     & 20k pairs & 77.8 & 69.2 & $\text{lr}=4\!\times\!10^{-6}$, $\beta=0.1$ \\
    IPO  (Ours)         & 20k pairs & 78.2 & 76.6 &  \\
    \bottomrule
  \end{tabular}
  \caption{Accuracy and head-to-head win rate against the SFT baseline for different preference-optimisation (PO) objectives. ``Uniform'' uses random sampling; ``w/o CRej'' applies hard-example mining; ``Ours'' adds counterfactual rationales.}
  \label{tab:po_methods}
\end{table*}
\footnotetext{Percentage of test questions where the PO model’s short answer is preferred over that of SFT.}

\begin{figure*}[!h]
\centering
\begin{minipage}{\textwidth}
\begin{lstlisting}[frame=single]
"split": "test",
"subtask": "region",
"idx": 2,
"subject_id": "10872780",
"study_id": "59144832",
"image_ids": ["661825e9-0b90c7bb-82d366fe-9d04a179-9c8e05ea_right lung"],
"question": "Can any presence of technical assessments be noted in the specified region?",
"answer": "Yes",
"answer_type": "closed",
"question_type": "presence",
"template": "Can any presence of ${category} be noted in the ${object}?",
"template_arguments": {
    "object": {"0": "right lung"},
    "category": {"0": "technical assessment"},
    "bbox": {"0": [85.71, 132.28, 302.85, 455.14]}}, ##All images are resized to 512x512
"rationale": "Yes. The image indicates that lung volumes are slightly low, which demonstrates a potential technical issue in the specified region. Additionally, the visualization of the specified region appears somewhat obscured, suggesting limitations in assessment in that area.",
"reports": ["Lung volumes are slightly low.  Heart size is mildly enlarged.  The aorta is slightly tortuous.  There is mild pulmonary edema with perihilar haziness and vascular indistinctness.  No focal consolidation, large pleural effusion or pneumothorax is present.  Assessment of the right apex is somewhat obscured by the patient's chin projecting over this area.  No acute osseous abnormality is detected.  There are degenerative changes noted involving both glenohumeral and acromioclavicular joints."]
\end{lstlisting}
\end{minipage}
\vspace{-0.3cm}
\caption{Example of Region-QA Data Attributes.}
\label{lst:data_attributes_example}
\end{figure*}

\begin{figure*}[!h]
\centering
\begin{minipage}{\textwidth}
\begin{lstlisting}[frame=single]
# X-Ray Diagnosis Explanation Generator
You are tasked with generating a concise radiological explanation for a chest X-ray diagnosis. Your explanation should simulate the diagnostic reasoning process of a professional radiologist evaluating the image directly.
## Input Format
<json>
{
  "question": "string - the original question asked",
  "answer": "string - the provided answer",
  "report": "string - the provided radiological findings"
}
</json>
## Output Format
{
  "explanation": "string - your generated explanation starting with the original answer"
}
## Requirements
1. Start your explanation by first stating the exact answer provided, and then follow it with your diagnostic explanation. 
2. Explain how specific imaging findings support the answer using standardized radiological language, avoid terms like 'reported,' 'mentioned,' noted,' 'stated,' or 'described'.
3. Include only imaging details directly relevant to the question and diagnosis.
4. Ensure explanation is based solely on image observations and presents a logical diagnostic process.

Based on the provided input (question, answer, and radiological findings), generate a radiological explanation following the requirements above. 

## Input:
{
question: {question},
answer: {answer},
report: {report},
}
## Output: {explanation}
\end{lstlisting}
\end{minipage}
\vspace{-0.3cm}
\caption{Prompt Template for Basic/Region-QA}
\label{lst:prompt_basic_region}
\end{figure*}

\begin{figure*}[!h]
\centering
\begin{minipage}{\textwidth}
\begin{lstlisting}[frame=single]
# Radiologist Image Comparison Explanation Generator
You are tasked with explaining radiological findings based on direct comparison of current and previous chest X-ray images.
Your explanation should simulate the diagnostic reasoning process of a professional radiologist evaluating the image directly, and should be generated based on a comparable question, its answer, and the radiological findings from both images.

## Input Format
<json>
{
  "question": "string - Clinical question about the difference or change between images",
  "answer": "string - Answer to the question",
  "study report": "string - Radiological findings from the current/most recent image examination",
  "reference report": "string - Radiological findings from the previous/comparison image examination"
}
</json>
## Output Format
{
  "explanation": "Your professional radiological explanation (starting with the answer, then explaining what you observe when comparing both images that supports this answer)"
}
## Requirements
1. Start your explanation by first stating the exact answer provided, and then follow it with your diagnostic explanation.
2. Explain how specific imaging findings support the answer using standardized radiological language, avoid terms like 'reported,' 'mentioned,' noted,' 'stated,' or 'described'.
3. Include only imaging details directly relevant to the question and diagnosis.
4. Include only imaging details directly relevant to the question and diagnosis. When showing changes, clearly attribute them to your visual comparison of both images.

Based on the provided input (question, answer, and radiological findings from both current and reference images), generate a professional radiological explanation following the requirements above. 
## Input:
{
  "question": {question},
  "answer": {answer} ,
  "study report": "{report_1}",
  "reference report": "{report_2}"
}
## Output: {explanation}
\end{lstlisting}
\end{minipage}
\vspace{-0.3cm}
\caption{Prompt Template for Comparison-QA}
\label{lst:prompt_compare}
\end{figure*}